\documentclass[11pt,a4paper]{article}
\usepackage[hyperref]{emnlp-ijcnlp-2019}
\usepackage{times}
\usepackage{latexsym}
\usepackage{amsmath}
\usepackage[inline]{enumitem}
\usepackage{arydshln}
\usepackage{url}
\usepackage{booktabs}
\usepackage{graphicx}
\usepackage{multirow}

\newcommand\shorten[1]{}

\aclfinalcopy 

\title{Partially-supervised Mention Detection}

\author{Lesly Miculicich\textsuperscript{$\dagger$ $\ddagger$} \ \ \  James Henderson\textsuperscript{$\dagger$ }\\ 
	\textsuperscript{$\dagger$ }{Idiap Research Institute, Switzerland}\\
	\textsuperscript{$\ddagger$}{\'{E}cole Polytechnique F\'{e}d\'{e}rale de Lausanne (EPFL), Switzerland}\\
	{\tt \{lmiculicich, jhenderson\}@idiap.ch}}

\date{}

\begin{document}
\maketitle
\begin{abstract}
  Learning to detect entity mentions without using syntactic information can be useful for integration and joint optimization with other tasks. However, it is common to have partially annotated data for this problem. Here, we investigate two approaches to deal with  partial annotation of mentions: weighted loss and soft-target classification. We also propose two neural mention detection approaches: a sequence tagging, and an exhaustive search. We evaluate our methods with coreference resolution as a downstream task, using multitask learning. The results show that the recall and F1 score improve for all methods.
\end{abstract}

\section{Introduction}
Mention detection is the task of identifying text spans referring to an entity: named, nominal or pronominal \cite{florian-etal-2004-statistical}. It is a fundamental component for several downstream tasks, such as coreference resolution \cite{soon-etal-2001-machine}, and relation extraction \cite{mintz-etal-2009-distant}; and it can help to maintain coherence in large text generation \cite{clark-etal-2018-neural}, and contextualized machine translation \cite{miculicich-werlen-popescu-belis-2017-using, miculicich-werlen-etal-2018-self, miculicich-etal-2018-document}.
 
Previous studies tackled mention detection jointly with named entity recognition \cite{xu-etal-2017-local, katiyar-cardie-2018-nested, ju-etal-2018-neural, wang-etal-2018-neural-transition}. There, only certain types of entities are considered (e.g. person, location, etc.); and the goal is to recognize mention spans and their types.
In this study, we are interested in the discovery of generic entity mentions which can potentially be referred to in the text, without the use of syntactic parsing information. 

\begin{figure}
	\center
	\includegraphics[width=\linewidth]{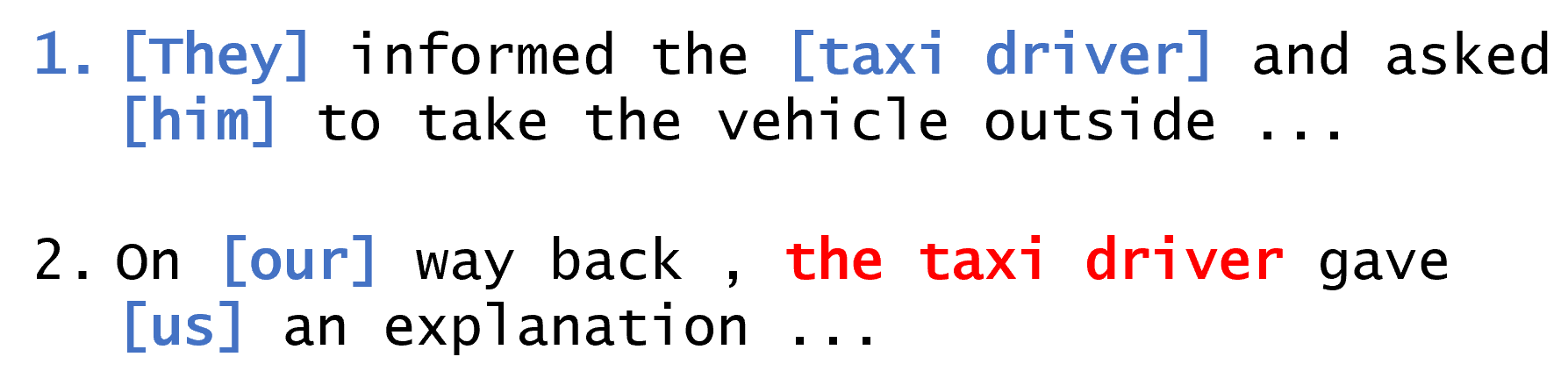}
	\caption{Samples from CoNLL 2012. Annotated mentions are within brackets}
	\vspace{-5mm}
	\label{fig:exa}
	
\end{figure}

Data from coreference resolution is suitable for our task, but the annotation is partial in that it contains only mentions that belong to a coreference chain, not single entity-mentions. 
Nevertheless, the missing mentions have approximately the same distribution as the annotated ones, so we can still learn this distribution from the data. Figure~\ref{fig:exa} shows an example from Ontonotes V.5 dataset \cite{W12-4501} where ``the taxi driver'' is annotated in sample 1 but not in 2.

Thus, we approach mention detection as a partially supervised problem, and investigate two simple techniques to compensate for the fact that some negative examples are true mentions: weighted loss functions, and soft-target classification. By doing this, the model is encouraged to predict more false-positive samples, so it can detect potential mentions which were not annotated.
We implement two neural mention detection methods:
a sequence tagging approach, and an exhaustive search approach.
We evaluate both techniques for coreference resolution by implementing a multi-task learning system. We show that the proposed techniques help the model to increase the recall significantly with a very small decrease in precision. In consequence, the F1 score of the mention detection and coreference resolution improves for both methods, and the exhaustive search approach yields a
significant improvement over the baseline coreference resolver. 

Our contributions are:
\begin{enumerate*}[label=(\alph*)]
 \item we investigate two techniques to deal with partially annotated data,
 \item we propose a sequence tagging method for mention detection that can model nested mentions, 
 \item we improve an exhaustive search method for mention detection, 
 \item we approach mention detection and coreferece resolution as multitask learning, and improve the recall of both tasks.
\end{enumerate*}

\section{Sequence tagging model}
Several studies have tackled mention detection and named entity recognition as a tagging problem. Some of them use one-to-one sequence tagging techniques \cite{lample-etal-2016-neural, xu-etal-2017-local}, while others use more elaborate techniques to include nested mentions \cite{katiyar-cardie-2018-nested, wang-etal-2018-neural-transition}. Here, we propose a simpler yet effective tagging approach that is able to manage nested mentions.

\begin{figure}
	\center
	\includegraphics[width=.8\linewidth]{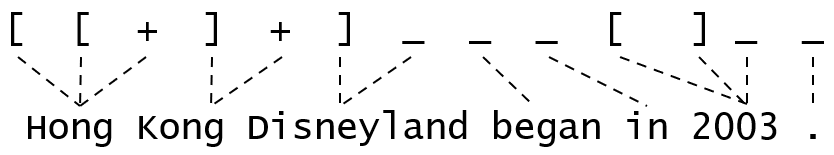}
	\caption{Tagged sentence example}
	\label{fig:tag}
	\vspace{-4mm}
\end{figure}

We use a sequence-to-sequence model which allows us to tag each word with multiple labels. The words are first encoded and contextualized using a recurrent neural network, and then a sequential decoder predicts the output tag sequence. During decoding, the model keeps a pointer into the encoder indicating the position of the word which is being tagged at each time step.
The tagging is done using the following set of symbols: $\{\textbf{[}, \textbf{]}, \textbf{+}, \textbf{-}\}$ . The brackets ``\textbf{[}'' and ``\textbf{]}'' indicate that the tagged word is the starting or ending of a mention respectively, the symbol ``\textbf{+}'' indicates that one or more mention brackets are open, and ``\textbf{-}'' indicates that none mention bracket is open. 
The pointer into the encoder moves to the next word only after predicting ``\textbf{+}'' or ``\textbf{-}''; otherwise it remains in the same position. 
Figure~\ref{fig:tag} shows a tagging example indicating the alignments of words with tags.

Given a corpus of sentences $X=(x_1,...,x_M)$, the goal is to find the parameters $\Theta$ which maximize the log likelihood of the corresponding tag sequences $Y=(y_1,...,y_T)$:
	\vspace{-1ex}
	\begin{equation}
          P_{\Theta}(Y|X) = \prod_{t{=}1}^T P_{\Theta}(y_t|X, y_1, ..., y_{t-1})
          .
	\vspace{-1ex}
	\end{equation}
The next tag probability is estimated with a softmax over the output vector of a neural network:
	\vspace{-3ex}
	\begin{align}\label{eq:dec}
	  P_{\Theta}(y_t|X, y_1, ..., y_{t-1}) &= softmax(o_t) \\
		o_t = relu(&W_o \cdot [ d_t, h_i ]+b_o)
	\vspace{-3ex}
	\end{align}
where $W_o, b_o$ are parameters of the network, $d_t$ is the vector representation of the tagged sequence at time-step $t$, modeled with a long-short term memory (LSTM)  \cite{hochreiter1997long}, and $h_i$ is the vector representation of the pointer's word at time $t$ contextualized with a bidirectional LSTM \cite{graves2005framewise}.
	\vspace{-1ex}
	\begin{align}
	  (h_1,...&,h_M) =BiLSTM(X) \\
	  d_t &= LSTM(y_1, ..., y_{t-1})
	\vspace{-1ex}
      \end{align}
        where the decoder is initialized with the last states of the bidirectional encoder, $d_0=h_{M}$. 

At decoding time, we use a beam search approach to obtain the sequence. The complexity of the model is linear with respect to the number of words. It can be parallelized at training time given that it uses ground-truth data for the conditioned variables. However, it cannot be parallelized during decoding because of its autoregressive nature.
\vspace{-4mm}
\section{Span scoring model}\label{sec:span}

Our span scoring model of mention detection is similar to the work of  \citet{D17-1018} for solving coreference resolution, and to \citet{ju-etal-2018-neural} for nested named mention detection, as both are exhaustive search methods. The objective is to score all possible spans $m_{ij}$ in a document, where $i$ and $j$ are the starting and ending word positions of the span in the document.  For this purpose, we minimize the binary cross-entropy with the labels $y$:
\nolinebreak
	\vspace{-1ex}
      \begin{align}
	{-}\frac{1}{M^2} \sum_{i=1}^M  \sum_{j=1}^M ( 
        &\, y_{m_{ij}} \!\ast log(P_{\Theta}(m_{ij}))
	\\[-2ex]
	&+ (1{-}y_{m_{ij}}) \!\ast log(1{-}P_{\Theta}(m_{ij})) \,)
	 \nonumber
	\vspace{-1ex}
      \end{align}
where $y_{m_{ij}} \in [0, 1]$ is one when there is a mention from position $i$ to $j$.  If $y_{m_{ij}}$ is zero when there is no mention annotated, then this is the same as maximising the log likelihood. But we will consider models where this is not the case.

The probability of detection is estimated as:
	\vspace{-1ex}
     \begin{align}\label{eq:sig}
	  P_{\Theta}(m_{ij}) &= \sigma(V \cdot relu(W_m  \cdot m_{ij} + b_m)) \\
	  m_{ij} = &relu( W_h   \cdot [h_i, h_j, \tilde{x}_{ij}]  + b_h)
	\vspace{-1ex}
	\end{align}
	where $V, W_m, W_h$ are weight parameters of the model, $b_m, b_h$ are biases, and $m_{ij}$ is a representation of the span from position $i$ to $j$. It is calculated with the contextualized representations of the starting and ending words $h_i, h_j$, and the average of the word embeddings $\tilde{x}_{ij}$:
	\vspace{-1ex}
	\begin{align}
	  (h_1,...,h_M) &=BiLSTM(X) \\
	  \tilde{x}_{ij} = &\dfrac{1}{j-i} \sum_{k=i}^{j} x_k 
	\vspace{-1ex}
	\end{align}
The complexity of this model is quadratic with respect to the number of words, however it can be parallelized at training and decoding time. \citet{D17-1018} uses an attention function over the embeddings instead of an average. That approach is less  memory efficient, and requires the maximum length of spans as a hyperparameter. Also, they include embeddings of the span lengths which are learned during training. As shown in the experimental part, these components do not improve the performance of our model.

\section{Partially annotated data}
The partial annotation of coreference data for mention detection means that spans which are not labelled as mentions may nontheless be true mentions of entities which could be referred to.  Thus, the approach of treating spans without mention annotations as true negative examples would be incorrect.  On the other hand, the ideal solution of sampling all possible mention annotations which are consistent with the given partial annotation would be intractable.  We want to modify the loss function of the model in such a way that, if the system predicts a false-positive, the loss is reduced.  This encourages the model to favor recall over precision by predicting more mention-like spans even when they are not labeled.  We assume that it is possible to learn the true mention distribution using the annotated mention samples by extrapolating to the non-annotated mentions, and propose two ways to encourage the model to do so.
\begin{description}[style=unboxed,leftmargin=0cm, itemsep=0pt, parsep=1pt,topsep=0pt, partopsep=0pt]
	\item[Weighted loss function:] We use a weighted loss function with weight $w{\in}]0,1[$ for negative examples only.
	The \emph{sequence tagging} model makes word-wise decisions, thus we consider words tagged as ``out of mention'', $y_t{=}``\textbf{-}\textquotedblright$, as negative examples, while the rest are positives. Although this simplification has the potential to increase inconsistencies, e.g.\ having non-ending or overlapping mentions, we observe that the LSMT-based model has the capability to capture the simple grammar of the tag labels with very few mistakes.  For  \emph{span scoring}, the distiction between negative and positive examples is clear given that the decisions are made for each span.
	\item[Soft-target classification:] Soft-targets allow us to have a distribution over all classes instead of having a single class annotation. Thus, we applied soft-targets to negative examples to reflect the probability that they could actually be positive ones. For  \emph{sequence tagging}, we set the target of negative examples,  $y_t{=}``\textbf{-}\textquotedblright$, to $(\rho, \rho, \rho, 1 - 3\rho)$ corresponding to the classes $( \textbf{[}, \textbf{]}, \textbf{+}, \textbf{-})$.
	For  \emph{span scoring}, we change the target of negative examples to $y_{neg}{=}\rho$. In both cases, $\rho$ is the probability of the example  being positive.
\end{description}



\section{Coreference Resolution}
We use multitask learning to train the mention detection together with coreference resolution. The weights to sum the loss functions of each task are estimated during training, as in \cite{cipolla2018multi}. 
The sentence encoder is shared and the output of mention detection serves as input to coreference resolution.
We use the coreference resolver proposed by \citet{D17-1018}. 
It uses a  pair-wise scoring function $s$ between a mention $m_k$ and each of its candidate antecedents $m_a$, defined as:
	\vspace{-1ex}
	\begin{equation*}
	s(m_k,m_a)= s_c(m_k,m_a) +  s_m(m_k) + s_m(m_a)
	\vspace{-1ex}
	\end{equation*}
where $s_c$ is a function which assesses whether two mentions refer to the same entity. 
We modified the mention detection score $s_m$.

For the \emph{sequence tagging} approach, the function $s_m$ serves as a bias value and it is calculated as:
	\vspace{-1ex}
  \begin{equation*}
    s_m = v. P(y_{t_i}=``\text{[}\textquotedblright)  . P(y_{t_j}=``\text{]}\textquotedblright)
	\vspace{-1ex}
  \end{equation*}
  where $y_{t_i}$ and $y_{t_j}$ are the labels of the first and last words of the span, and $v$ is a scalar parameter learned during training.  At test time, only mentions in the one-best output of the mention detection model are candidate mentions for the coreference resolver.  During training, the set of candidate mentions includes both the spans detected by the mention detection model and the ground truth mentions.  The mention decoder is run for one pass with ground-truth  labels in the conditional part of the probability function (Eq.~\ref{eq:dec}), to get the mention detection loss, and run for a second pass with predicted labels to provide input for the coreferece task and compute the coreference loss.

For the \emph{span scoring} approach, $s_m$ is a function of the probability defined in Eq.~\ref{eq:sig}, scaled by a parameter $v$ learned during training. 
	\vspace{-1ex}
 \begin{equation*}
   s_m = v.  P(m_{i,j})
	\vspace{-1ex}
  \end{equation*}
Instead of the end-to-end objective of \citet{D17-1018}, we use a multitask objective which adds the loss function of mention detection. Also, we do not prune mentions with a maximum length, nor impose any maximum number of mentions per document.  We use the probability of the mention detector with a threshold $\tau$ for pruning.
	

\section{Experiments}
We evaluate our model on the English OntoNotes set from the CoNLL 2012 shared-task \cite{W12-4501}, which has 2802 documents for training, 343 for development, and 348 for testing. 
The setup is the same as \citet{D17-1018} for comparison purposes, with the hyper-parameters $\rho, w, \tau$ optimized on the development set. We use the average F1 score as defined in the shared-task \cite{W12-4501} for evaluation of mention detection and coreference resolution.

	\begin{table}
	\center
	\small
	\begin{tabular}{l c c c   } \toprule
		Model  & Rec. & Prec. & F1   \\ \hline
		Sequence tagging &  73.7 & 77.5 & 75.6    \\
		Span scoring  & 72.7 & 79.2 & 75.8  \\
		+ span size emb. &  71.6 & 80.1 & 75.6   \\
		\,  - avg. emb. + att. emb.  &   72.1 & 78.9 & 75.4 \\  		
		\bottomrule
	\end{tabular}    
	\caption{Mention detection evaluation} 
	\vspace{-4mm}
	\label{tab:ment}
\end{table}

\subsection{Mention detection}
First, we evaluate our stand-alone mention detectors.  For this evaluation, all unannotated mentions are treated as negative examples. Table~\ref{tab:ment}  show the results on the test set with models selected using the best F1 score with $\tau{=}0.5$, on the development set. We can see that \emph{sequence tagging}  performs almost as well as \emph{span scoring} in F1 score, even though the latter is an exhaustive search method. We also evaluate the \emph{span scoring} model with different components from \citet{D17-1018}. By adding the span size vector, the precision increases but the recall decreases.  Replacing the average embedding $\tilde{x}$  with attention over the embeddings requires a limited span size for memory efficiency, reslting in decreased performance.

\subsection{Coreference Resolution}
Table~\ref{tab:res} shows the results obtained for our multi-task systems for coreference resolution and mention detection with and without the loss modification.  
The \emph{sequence tagging} method obtains lower performance compared to \emph{span scoring}. This can be attributed to its one-best method to select mentions, in contrast to \emph{span scoring} where uncertainty is fully integrated with the coreference system. The \emph{span scoring} method performs similarly to the coreference resolution baseline, showing that the naive introduction of a loss for mention detection does not improve performance (although we find it does decrease convergence time).
However, adding the modified mention loss does improve coreference performance. For \emph{sequence tagging}, the weighted loss results in higher performance, while for the \emph{span scoring} soft-targets work best. 
In both cases, the recall increases with a small decrease in precision, which improves the F1 score of mention detection, and  in turn improves conference resolution.

	\begin{table}
	\center
	\small
	{\def\arraystretch{1.1}\tabcolsep=4pt
		\begin{tabular}{l c c c | c  } \toprule
			& \multicolumn{3}{c|}{Mention} & Coref. \\ 
			Model & Rec. & Prec. & F1 & Avg. F1 \\ \hline 
			\citet{D17-1018} & --& -- & -- & 67.2  \\ \hline 
			Sequence tagging  & 73.1 & 84.9 & 78.6 & 59.9 \\ 
			+ wt. loss $w{=}0.01$ & 77.3 & 83.2 & 80.1 & 64.1 \\
			+ soft-target $\rho{=}0.1$ & 74.3 & 84.0 & 78.8 & 61.2 \\ 
			Span scoring & 75.3 & 88.3 & 81.3 & 67.0 \\ 
			+ wt. loss $w{=}0.3$ & 76.3 & 88.1 & 81.8 & 67.1 \\
			+ soft-target $\rho{=}0.1$& 78.4 & 87.9 & 82.9 & 67.6 \\ \bottomrule 
		\end{tabular}   
		\caption{Coreference resolution evaluation}
		\vspace{-2mm}
		\label{tab:res}
	}		
	\end{table}

\begin{figure}[t]
	\center
	\includegraphics[width=\linewidth]{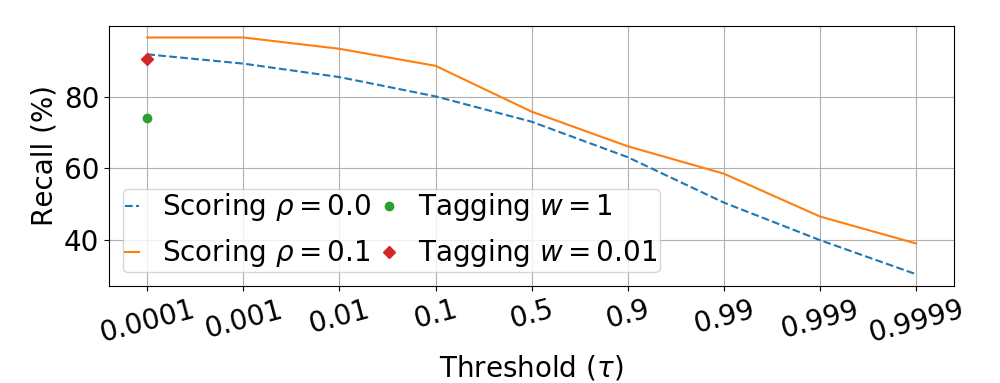}
	\caption{Recall comparison}
	\label{fig:recall}
	\vspace{-4mm}
\end{figure}

\subsection{Recall performance}
Figure~\ref{fig:recall} shows a comparison of the mention detection methods in terms of recall. The unmodified \emph{sequence tagging} model achieves 73.7\% recall, and by introducing a weighted loss at $w{=}0.01$, it reaches 90.5\%. The lines show the variation of recall for the \emph{span scoring} method with respect to the detection threshold $\tau$. The dotted line represents the unmodified model, while the continuous line represents the model with soft-targets at $\rho{=}0.1$, which shows higher recall for every $\tau$.

\section{Conclusions}
We investigate two simple techniques to deal with partially annotated data for mention detection, and propose two methods for mentions detection. We evaluate on coreference and mention detection with a multitask learning approach, and show that the techniques effectively increase the recall of mentions and mention coreference with a small decrease in precision, which overall improves the F1 score and improves coreference. In future, we plan to use these methods to maintain coherence over long distances when reading, translating and generating very large texts, by keeping track of abstract representations of entities.

\bibliographystyle{acl_natbib}
\bibliography{mention_detection}

\end{document}